\documentclass[runningheads]{llncs}
\usepackage[T1]{fontenc}
\usepackage{graphicx}
\usepackage{amsmath}
\usepackage{booktabs}
\usepackage{multirow}
\usepackage[table]{xcolor}
\usepackage{orcidlink}
\begin{document}
\title{ICPR 2024 Competition on Rider Intention Prediction}
\author{Shankar Gangisetty\inst{1}$^*$\orcidlink{0000-0003-4448-5794} \and
Abdul Wasi\inst{1}$^*$\orcidlink{0000-0003-4465-7066} \and
Shyam Nandan Rai\inst{2}\orcidlink{0000-0002-7448-2271} \and C. V. Jawahar\inst{1}\orcidlink{0000-0001-6767-7057} \and Sajay Raj\inst{3} \and Manish Prajapati\inst{4} \and Ayesha Choudhary\inst{4} \and Aaryadev Chandra\inst{5} \and Dev Chandan\inst{5} \and Shireen Chand\inst{5} \and Suvaditya Mukherjee\inst{5}
}

\authorrunning{S. ~Gangisetty et al.}
\institute{IIIT Hyderabad, India \and Politecnico di Torino, Italy 
\and Woxsen University, Hyderabad \and Jawaharlal Nehru University, New Delhi \and NMIMS, Mumbai  
}
\maketitle              % typeset the header of the contribution
\def\thefootnote{*}\footnotetext{equal contribution}

\begin{figure}[th]
  \centering
  \includegraphics[width=1\linewidth]{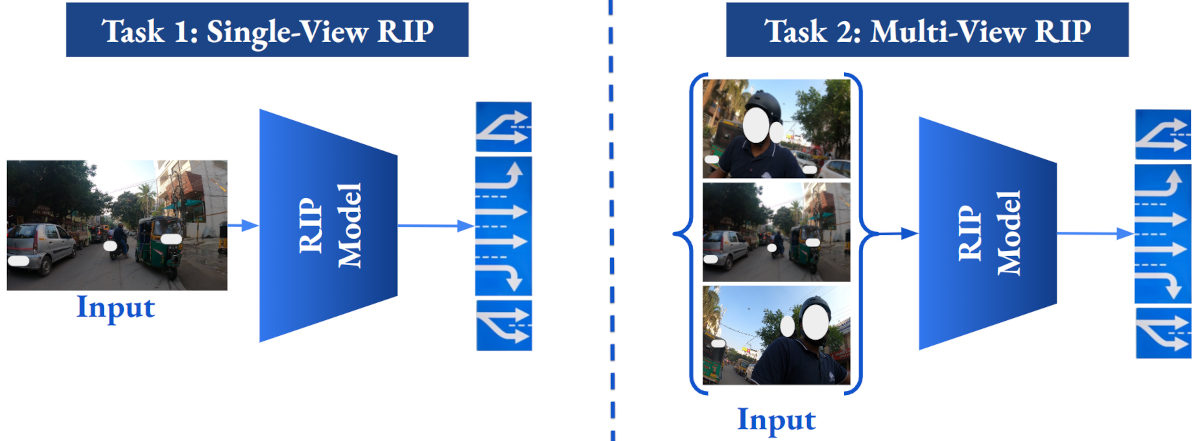} 
  \caption{\textbf{Overview}: We propose a \textit{Rider Intent Prediction (RIP)} challenge aimed at enhancing the safety of two-wheeler riders by introducing two tasks in our challenge: single-view (i.e., frontal-view) RIP (Task 1) and multi-view RIP (Task 2), where, we curate a new dataset and benchmarked the tasks performances.}
\label{fig:teasure_rip}
\end{figure}

%\begin{figure}[th]
%  \centering
%  \includegraphics[width=1\linewidth]{Images/accident_stats_india_v2.png} 
%  \caption{\textbf{Road Accidents Statistics:}  According to sources (Frontline, OpenCity, MoRTH~\cite{thirdcite}), India is particularly susceptible to motorbike accidents.}
%\label{fig:main_road_accidents}
%\end{figure}

%\textbf{\textcolor{magenta}{Note: We appreciate the reviewers for their valuable feedback. We addressed the concerns of reviewers which are highlighted in} \textcolor{blue}{blue text}.}

\begin{abstract}
The recent surge in the vehicle market has led to an alarming increase in road accidents. This underscores the critical importance of enhancing road safety measures, particularly for vulnerable road users like motorcyclists. Hence, we introduce the rider intention prediction (RIP) competition that aims to address challenges in rider safety by proactively predicting maneuvers before they occur, thereby strengthening rider safety. This capability enables the riders to react to the potential incorrect maneuvers flagged by advanced driver assistance systems (ADAS). We collect a new dataset, namely, rider action anticipation dataset (RAAD) for the competition consisting of two tasks: single-view RIP and multi-view RIP. The dataset incorporates a spectrum of traffic conditions and challenging navigational maneuvers on roads with varying lighting conditions. 
For the competition, we received seventy-five registrations and five team submissions for inference of which we compared the methods of the top three performing teams on both the RIP tasks: one state-space model (Mamba2) and two learning-based approaches (SVM and CNN-LSTM). The results indicate that the state-space model outperformed the other methods across the entire dataset, providing a balanced performance across maneuver classes. The SVM-based RIP method showed the second-best performance when using random sampling and SMOTE. However, the CNN-LSTM method underperformed, primarily due to class imbalance issues, particularly struggling with minority classes. This paper details the proposed RAAD dataset and provides a summary of the submissions for the RIP 2024 competition.


\end{abstract}
\section{Introduction}
\label{sec:intro}
The Global Autonomous Vehicle market is projected to reach $2,162$ Billion by $2030$~\cite{firstcite}. However, around $1.3$ Million lives are lost in road accidents every year. A considerable proportion of road accidents are attributed to the behavior of vulnerable agents like motorbike riders and pedestrians.
%as shown in Fig.~\ref{fig:main_road_accidents}. 
India has the highest number of two-wheeler riders in the world with a significant rise each year~\cite{secondcite}. According to a report published by the Ministry of Road Transport and Highways (MoRTH)~\cite{thirdcite}, the number of fatal accidents involving two-wheeler riders has consistently increased over the past few years. Furthermore, in congested traffic situations and on roads lacking a clear structure, the potential for accidents persists when a motorcyclist plans to maneuver, regardless of their skill level. For instance, motorcycle riders frequently engage in hazardous actions such as crossing the road diagonally without signaling or navigating through tight spaces between other vehicles to evade congestion. Existing ADAS
systems detect critical maneuvers only after the driver has initiated them ~\cite{Gebert19}, leading to the onset of a hazardous situation.

\begin{figure}[th]
  \centering
  \includegraphics[width=1\linewidth]{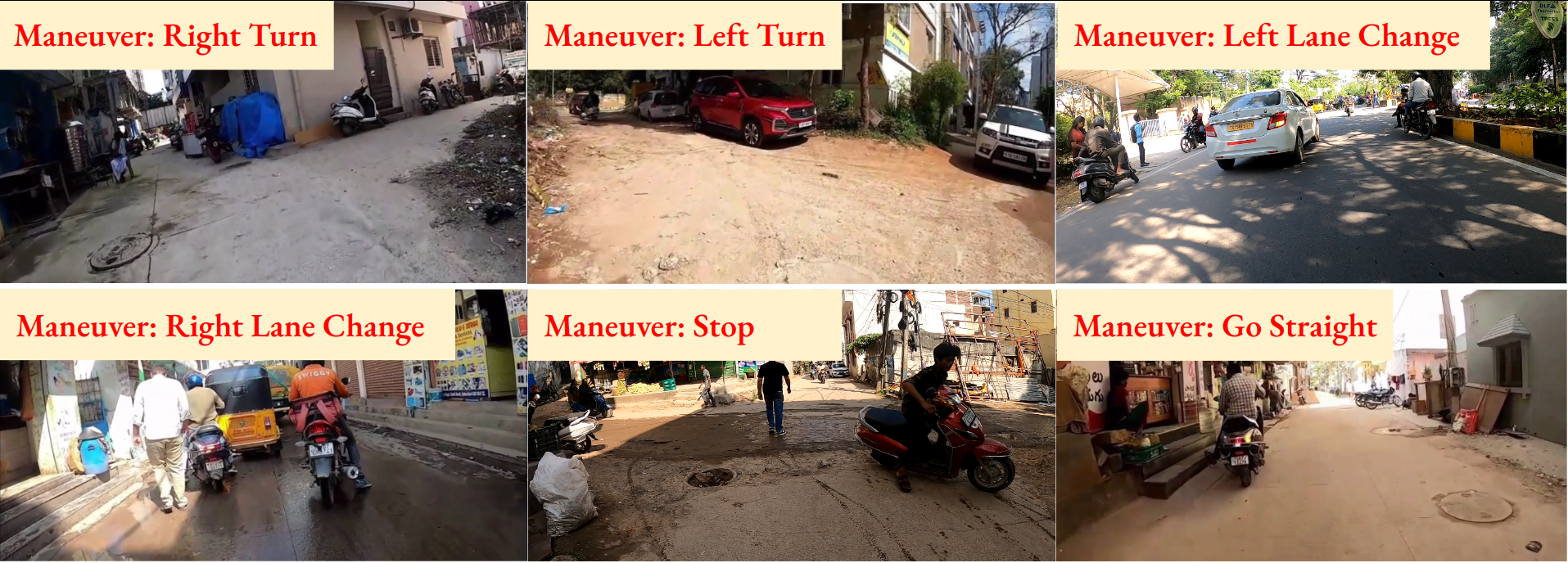} 
  \caption{The RAAD dataset illustrates diverse two-wheeler rider maneuvers on unstructured road conditions.}
\label{fig:main_rip}
\end{figure}

We address these challenges to ensure better rider safety by proposing a unique RIP competition to anticipate the rider maneuver prior before the actual maneuver occurs. The vehicle types involved in the RIP competition data capture include two-wheelers such as motorbikes and scooters. The behaviors (i.e., rider maneuvers) predicted by RIP tasks include \textit{left turn}, \textit{right turn}, \textit{left lane change}, \textit{right lane change}, \textit{straight}, and \textit{slow-stop} on the roads as shown in Fig.~\ref{fig:main_rip}. With this context, we introduced the following tasks for the competition: 
\begin{itemize}
    \item {\bf Task 1:} Single-view (i.e., frontal-view) rider intention prediction 
    \item {\bf Task 2:} Multi-view rider  intention prediction
\end{itemize}

The closest attempt to address the two-wheeler RIP challenge is found in four-wheeler driver intention prediction (DIP)~\cite{Gebert19,Jain15,Jain16,Khairdoost20}, which aims to predict the maneuver during its execution. To solve the DIP task there are a handful of DIP datasets being studied by groups from Stanford and Cornell University (Brain4Cars)~\cite{jain2016brain4cars}, Honda Research Institute (HDD)~\cite{ramanishka2018hdd}, and Australian National University and  EPFL (Viena$^2$)~\cite{aliakbarian2018viena2}. Brain4Cars~\cite{jain2016brain4cars} and HDD~\cite{ramanishka2018hdd} datasets are captured in the USA. Here the roads are properly laid, the traffic and pedestrian density are scarce, and the traffic rules are not flouted. In Viena$^2$~\cite{aliakbarian2018viena2}, the data is captured in a similar setup, however, using a simulator.  
Moreover, none of these datasets are captured using a two-wheeler, where the maneuvers are more complicated, especially in dense and unstructured roads of third-world countries. This makes RIP a unique task of great practical importance in Indian traffic scenarios and similar countries lacking road infrastructure.

%The competition is proposed to boost the research in Indian rider maneuvers to reduce the high fatality of road accidents, which needs alarming attention.
On the other hand, there are no RIP methods to address these challenges. To accomplish these shortcomings, we introduce a Rider Action Anticipation Dataset (RAAD) comprising
multiple view videos i.e., left mirror-view, frontal-view, and right mirror-view with sequences during the execution of rider maneuvers to obtain surrounding traffic context. This dataset contains video clips captured across a range of lighting conditions and traffic densities. Furthermore, it includes footage from diverse interior roads inaccessible via conventional four-wheeled vehicles. With this new RAAD dataset (see Fig.~\ref{fig:main_rip}), we aim to compare and develop intention prediction techniques for frontal-view and multi-view rider maneuvers.
%We compare different methods on the proposed RAAD benchmark. 
The contributions are summarized as:
\begin{itemize}
    %\item We provide a unique classification-based RAAD dataset from an unstructured two-wheeler driving environment. There are no other open rider maneuver video datasets released for our purpose.
    \item We introduce a novel classification-based RAAD dataset, specifically curated from an unstructured two-wheeler driving environment. To our knowledge, no other open rider maneuver video datasets have been released for similar purposes.
\item We evaluated various methods submitted by participating teams on RAAD dataset, contributing to solutions for frontal and multi-view RIP tasks.
\item We present quantitative benchmark results for the RAAD dataset across different methods of participating teams, and showcase that Mamba2 state-space models excel in rider intention prediction.
\end{itemize}

\noindent \textbf{Organization.}
This competition consisted of two tasks: the single-view (i.e., frontal-view and multi-view RIP tasks (see Section~\ref{sec:task_eval}). The focus of the competition was on rider intention prediction, with the goal of developing robust methods to enhance rider safety by predicting maneuvers such as \textit{left turn}, \textit{right turn}, \textit{left lane change}, \textit{right lane change}, \textit{straight}, and \textit{slow-stop}. The RAAD dataset curated for this challenge is detailed in Section~\ref{sec:dataset}. The competition was hosted on a web portal \url{https://mobility.iiit.ac.in/icpr_2024_rip/}, which facilitated participant interaction, provided challenge information, registration links, schedules, download links, online submissions, and real-time leaderboard updates. A total of \textbf{seventy-five} registrations were received, with \textbf{five teams} making one or more submissions after the test set was released five days before the submission deadline. Participants were allowed a maximum of three submissions for both Task 1 and Task 2. For each task, only the highest-performing result was considered, and the overall score was calculated as described in Section~\ref{sec:task_eval}. The top three solutions for both Task 1 and Task 2 were recognized as winners and awarded first, second, and third place, respectively.

\begin{figure}[th]
  \centering
  \includegraphics[width=1\linewidth]{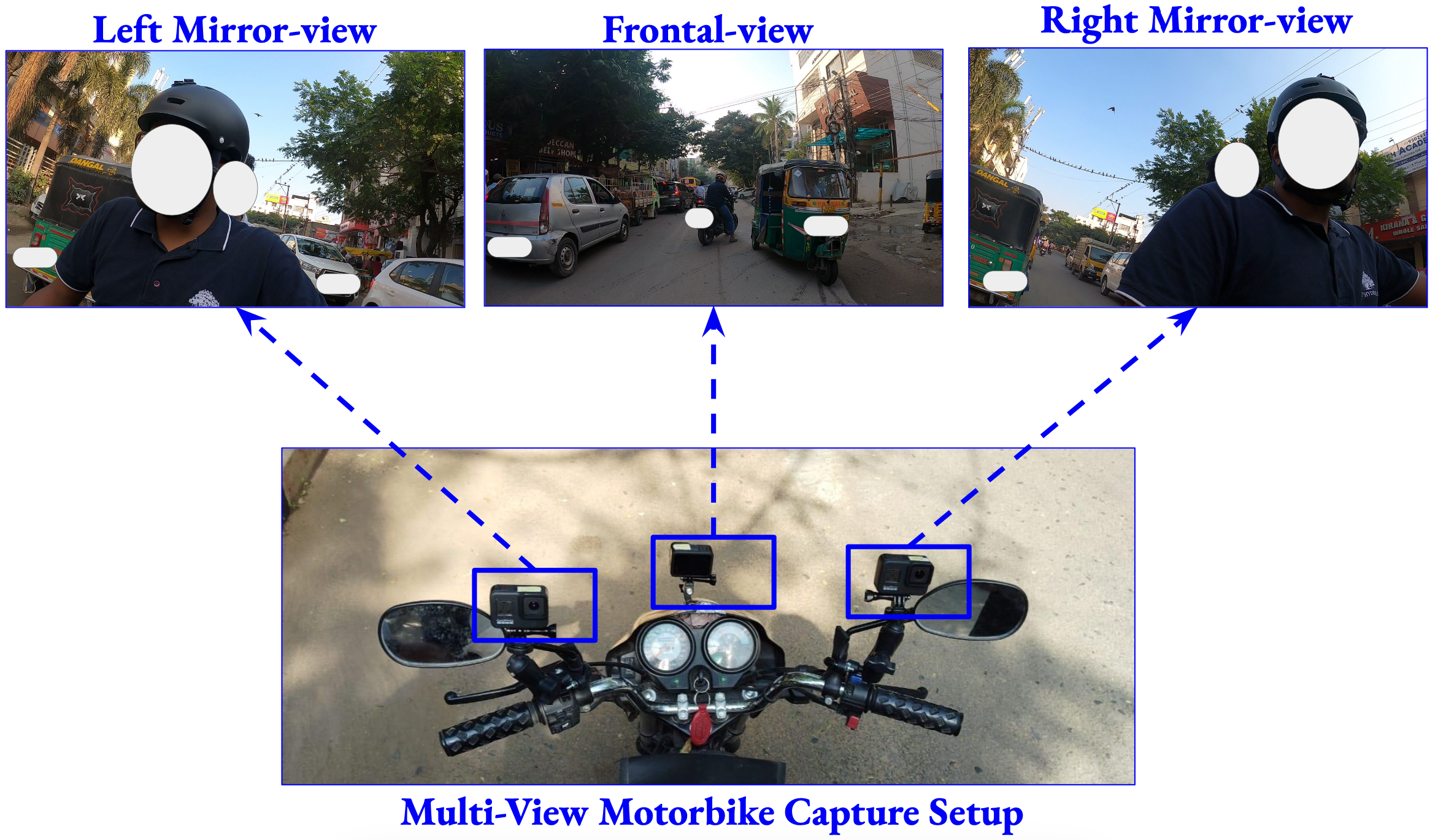} 
  \caption{\textbf{Data capture setup.} Three cameras are oriented towards frontal, left mirror, and right mirror views.}
\label{fig:motorbike_setup}
\end{figure} 

\section{RAAD Benchmark}
\label{sec:dataset}
In this section, we introduce the RAAD dataset. We outline the data capture setup, present the annotation process, and analyze the dataset statistics.

\subsubsection{Data Capture Setup.}
To create the RAAD dataset, the cameras on two-wheeler were arranged as shown in Fig.~\ref{fig:motorbike_setup}. 
The two-wheeler was equipped with three monocular (GoPro $8$) cameras with $1080$p resolution. These cameras are strategically positioned to capture frontal, left mirror, and right mirror views, providing a comprehensive overview of the surrounding traffic. This setup is crucial for accurately anticipating maneuvers in real-time.
We then curate a diverse dataset of two-wheeler rider intentions, encompassing various maneuvers on the roads as shown in Fig.~\ref{fig:main_rip}. This dataset comprises rider maneuvers such as \textit{left turn}, \textit{right turn}, \textit{left lane change}, \textit{right lane change}, \textit{straight}, and \textit{slow-stop}. Each video clip in the dataset consists of a single maneuver capturing the video streams at a rate of $30$ frames per second. 

\begin{figure*}[t]
  \centering
  \includegraphics[width=0.8\linewidth]{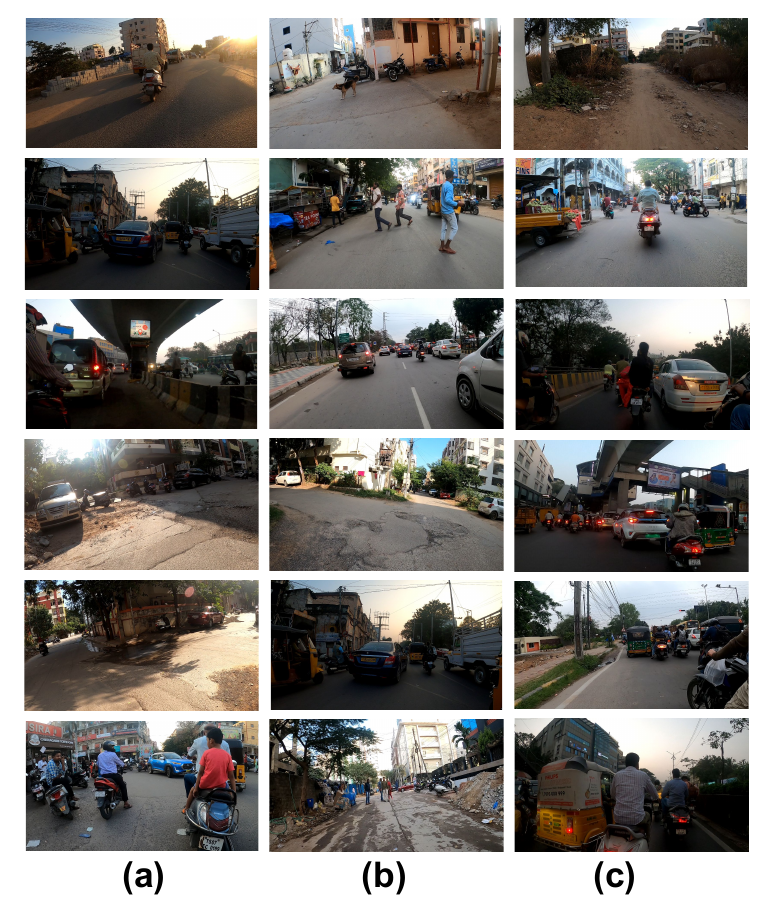}
  \caption{\textbf{RAAD data samples.} Shows images captured in diverse (a) lighting conditions, (b) route type, and (c) traffic density.}
  \label{fig:diverse-riding}\vspace{-5mm}
\end{figure*}

\subsubsection{Data Collection.} The dataset was collected for $50$ hours over $700$ kms on $12$ different routes, spanning over $6$ months from Nov'$23$ to April'$24$. The rider maneuvering dataset is diverse, encompassing $12$ riders with varying experience levels (ranging from $5$ to $30$ years). It includes data collected from a variety of environments such as residential areas, markets, rural and suburban streets, as well as narrow interior roads where three-wheelers or four-wheelers cannot traverse. The data spans different times of day and night, various weather conditions, and varying traffic densities, providing a comprehensive overview of real-world riding scenarios.
%Table~\ref{tab:comparisoin_table} compares RAAD against earlier riding datasets. 
In Fig.~\ref{fig:diverse-riding}, RAAD dataset highlights the variety of maneuvers in unstructured riding conditions and demonstrates the broad diversity captured.

\subsubsection{Annotation Process.}
We annotate different rider intentions by categorizing them based on the type of maneuver. Here, the labels used are \textit{straight} (ST), \textit{right turn} (RT), \textit{left turn} (LT), \textit{right lane change} (RLC), \textit{left lane change} (LLC), and \textit{slow/stop} (SS). We selected these labels of maneuvers based on two factors. Firstly, they are frequently observed on Asian roads. Secondly, they are commonly observed without recurring interference from the surrounding traffic context (i.e., other traffic agents are not acting as a cause for the maneuver, for example by an obstruction, yield, cut-in, etc).
%This approach ensures the inclusion of a wide range of traffic scenarios that are encountered across roads. 

\subsubsection{Data Statistics.}
\begin{figure*}[t]
  \centering
  \includegraphics[width=1\linewidth]{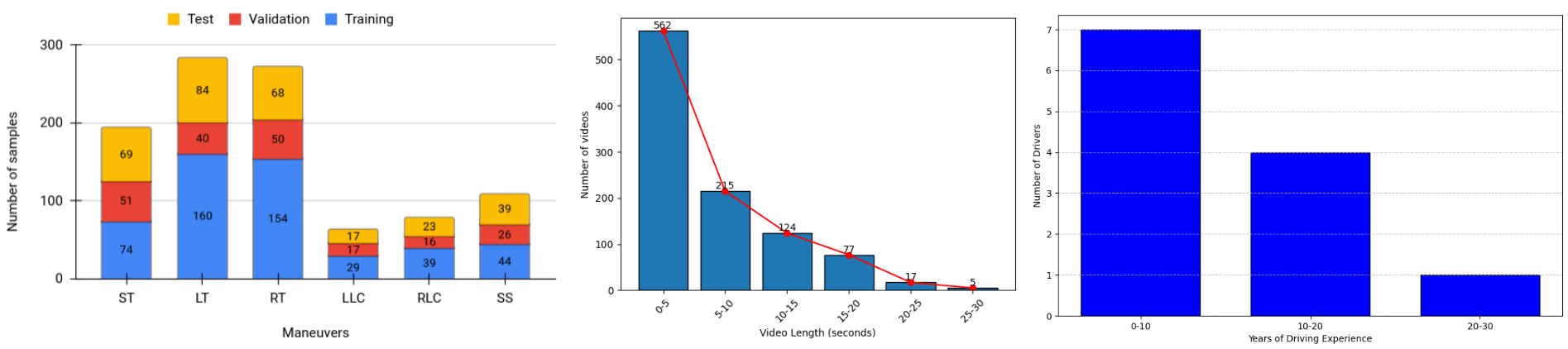}
  \caption{\textbf{Data statistics.} \textit{Left:} Number of videos for each maneuver. \textit{Middle:} Average video length of our dataset. \textit{Right:} Years of driving experience.}
  \label{fig:data_stats}\vspace{-5mm}
\end{figure*}

%\vspace{-.2cm}
In Fig.~\ref{fig:data_stats}, we provide a detailed overview of the dataset statistics. The dataset comprises a total of $1,000$ video samples, with a varying length of $5$ to $30$ seconds. Each sample comprises three video clips, each from a different camera view. For every sample, participants are provided with corresponding labels and video embedding feature information. The provided data includes three categories of feature embeddings including VGG-16~\cite{vgg}, ResNet-50~\cite{resnet}, and R(2+1)D~\cite{r21d}, offering participants a comprehensive set of information for the competition. The dataset is split into training ($50$\%), validation ($20$\%), and testing ($30$\%) sets. 
%This division is performed without considering held-out subjects, acknowledging the inherent data imbalance stemming from the naturalistic nature of the dataset. 

\section{Tasks and Evaluation}
\label{sec:task_eval}
In this section, we discuss the RIP tasks as illustrated in Fig.~\ref{fig:teasure_rip} and explain the evaluation metrics.

\subsubsection{Task 1: Frontal-view RIP}
Given the RAAD dataset containing $1,000$ frontal-view videos of driving maneuvers, the objective of this task is to predict the rider's intention before the actual maneuver begins. The dataset is divided into a training set with $500$ videos and a validation set with $200$ videos, both accompanied by extracted feature embeddings and maneuver label annotations. Three types of embeddings, namely, VGG-16, ResNet-50, and R(2+1)D are provided. The test set, consisting of $300$ videos, is also available but without maneuver label annotations.
%Given the RAAD dataset with $1,000$ frontal-view videos of driving maneuvers, the goal of this task is to predict the rider intention a few seconds before the actual maneuver occurs. The training set (with $500$ videos) and the validation set (with $200$ videos) are provided with extracted feature embedding and maneuver label annotations. Three categories of embedding, namely, VGG-16, ResNet-50, and R(2+1)D are provided. The test set (with $300$ videos) are also provided without maneuver label annotations.

\subsubsection{Task 2: Multi View RIP}
In multi-view RIP, the objective remains the same as in Task 1, but it incorporates the multi-view exterior traffic context. This includes $1,000$ videos capturing frontal views, left mirror-views, and right mirror-views of driving maneuvers.
%In multi-view RIP, the goal remains same as task 1 except that we consider multi-view exterior traffic context, encompassing $1,000$ frontal-view, left side-mirror view, and right side-mirror view videos of driving maneuvers.

\subsubsection{Evaluation Metrics.}
We evaluate RIP with two metrics: the classification accuracy, handling the driving maneuvers as classes, and the $F_1$ score for detecting the maneuvers (i.e. the harmonic mean of the precision and recall)
The accuracy metric is defined as follows:
\begin{equation}
    Acc := 1/n \sum_{i=1}^{n}\sigma(p(s_{i}),t_{i})
\end{equation}
with $\sigma(i,j):= 1$ (\text{if } $i = j$) \text{ and } $0$ (\text{otherwise})

where $p$ refers to the prediction of the classifier for the
sample $s_i$ with corresponding target label $t_i$ and $n$ the total number of samples in the data set. Here, each sample $s_i$ is a video of a specific maneuver.

Another metric that we use for measuring the performance of the model is $F_1$ score.
Using precision and recall, we calculate the $F_1$ score as
\begin{equation}
    F_1 = \frac{2 . P . R}{P + R}
\end{equation}
where P and R are,
\begin{equation}
    P = \frac{tp}{tp + fp + fpp}, R = \frac{tp}{tp + fp + mp}
\end{equation}

Precision and recall are defined as follows:
\begin{itemize}
    \item true prediction ($tp$): correct prediction of the maneuver in a video.
    \item false prediction ($fp$): prediction is different than the
actual performed maneuver
    \item false positive prediction ($fpp$): a maneuver-action predicted, but the driver is driving straight
    \item missed prediction ($mp$): a driving-straight predicted, but a maneuver is performed
\end{itemize}

\section{Submitted Methods}
\label{sec:methods}
Participants were allowed to submit up to three entries, with only their best-performing submission being considered as the official entry for the competition. In total, $7$ submissions were received from $5$ teams. Table~\ref{tab:baselines} shows the teams and the results of the top submissions from these teams. The following section provides details on the methods adopted by the top three teams, along with a description of the baseline established at the time of the dataset release.

\subsection{Baseline}
\label{sec:baseline}
Our baseline is an RNN based classification model for frontal-view RIP (task 1) which takes extracted features from VGG-16~\cite{vgg}. 
For multi-view RIP (task 2), we provide a baseline that does concatenation (early fusion) of features from multiple views and then follows the same procedure as task 1.
We provided the baseline code\footnote{https://github.com/wasilone11/ICPR-RIP-2024} and extracted features (i.e., VGG-16~\cite{vgg}, ResNet-50~\cite{resnet} and R(2+1)D~\cite{r21d}) in order to support participants with limited computing facilities. The participants can use any one of these three feature embedding in solving the tasks.  

\begin{figure*}[t]
  \centering
  \includegraphics[width=1\linewidth]{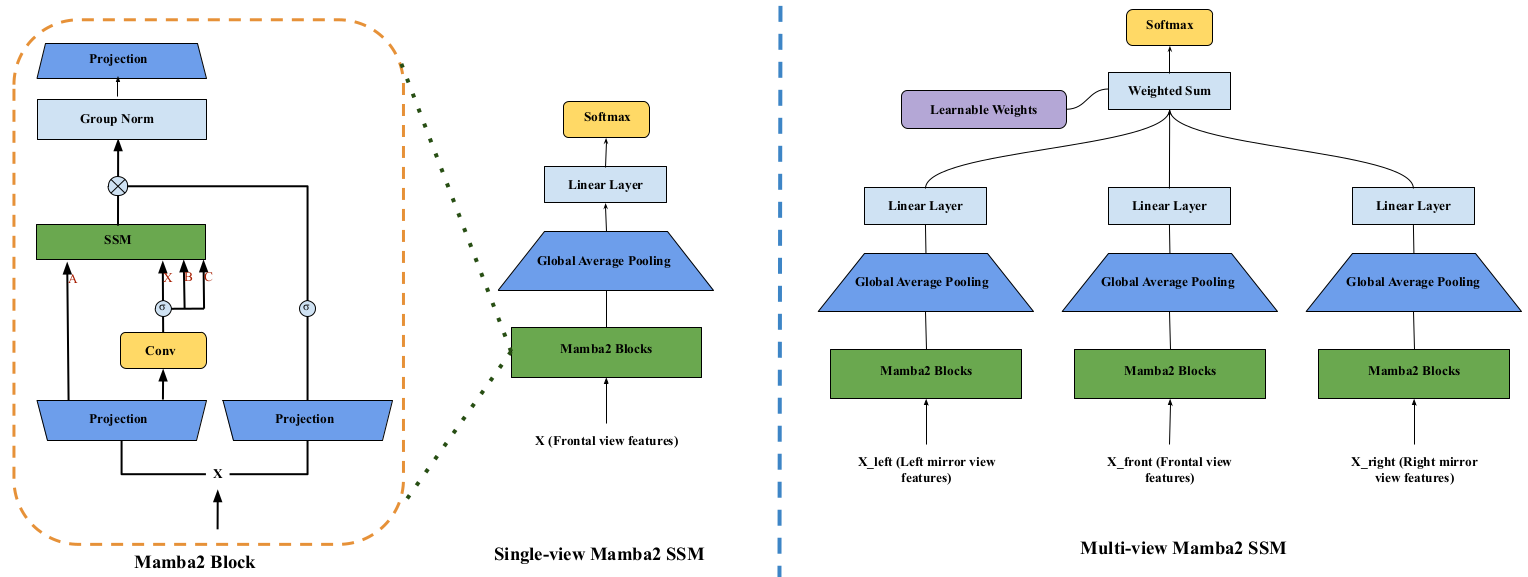}
  \caption{\textbf{Mamba2 SSM method.} The Mamba2 block, single-view and multi-view Mamba2 SSM methods for tasks.}
  \label{fig:ssm}\vspace{-5mm}
\end{figure*}

\subsection{\textit{Team Cisco:} Mamba2 state-space model (SSM) method}
\label{sec:ssm}
This method is contributed by Sajay Raj from Woxsen University, Hyderabad, and is associated with team \textit{Cisco}.
The task of RIP could be
considered a task of high-dimensional time
series classification, where subtle changes in
frame embeddings throughout video clips could
hint towards the prediction class, thus a
recurrent model that’s meant for high
dimensional and information dense data is
appropriate for the task like Mamba2~\cite{Dao2024ssm}.
Mamba2 is a new state-space duality architecture built for information-dense data.
They are, by design, fully recurrent models that
make them suitable for operating on long
sequences while being able to selectively
prioritize information with the data-driven
selection mechanism. They achieve performance
equal or better than transformers on sequence
modeling tasks while maintaining linear-time
complexity.

\subsubsection{Mamba2 block.} The Mamba2~\cite{Dao2024ssm} block used in our method is illustrated in Fig.~\ref{fig:ssm}. The input sequence $X$ (typically of shape $(B,V,D)$, where $B$ is batch size, $V$ is sequence length, and $D$ is input dimension) is linearly projected to create two intermediate representations, $A$ and $Z$. Then a $1D$ convolutional operation is applied to $A$,
followed by a nonlinear SiLU activation, producing $X,B,C$. The SSM processes the input using the dynamically generated parameters $X$, $B$, and $C$ with $A$. The outputs from the SSM path and the skip
connection path containing $Z$ are combined and normalized with a layer norm. Finally, the
output is projected back to the
original input dimension, completing the
block's operation.

\subsubsection{Frontal-view Mamba2 SSM Method.}
Our frontal view method for RIP uses the Mamba2 blocks to process temporal sequences of frame
embeddings extracted from video clips as shown in Fig.~\ref{fig:ssm}. This
the approach allows us to capture both short-term
and long-term dependencies in the rider's
behavior, which is crucial for the accurate intention
prediction.
The input to our method consists of VGG-16
frontal view features of dimension 512. 
To handle the variable-length nature of video
clips, we employ padding within each batch
during training and inference. This approach
allows us to efficiently process sequences of
different lengths without losing temporal
information or introducing unnecessary
computational overhead.
The architecture of our frontal view method is
designed to progressively refine the temporal
representation of the input sequence. It consists
of a series of Mamba2 blocks, each of which
processes the entire sequence and updates its
internal state. We then apply the global average
pooling across the temporal dimension to condense 
into a fixed-size representation for classification tasks. The pooled features are then passed through a
final linear layer, which maps them to the class
probabilities corresponding to different rider
maneuvers and a softmax activation function to
ensure that the output represents a valid probability distribution over the possible
maneuver classes.

\subsubsection{Multi-view Mamba2 SSM Method.}
We built a multi-view RIP based on an
ensemble of three frontal view Mamba2 method as shown in Fig.~\ref{fig:ssm},
each dedicated to processing the features from
frontal, left, and right views. This design choice
allows each model to specialize in extracting
relevant information from its respective view
while maintaining the ability to capture
view-specific temporal dynamics.
A learnable weighting mechanism is used to combine each model's predictions. The final prediction is computed as a weighted sum of each frontal view method output. To ensure that the resulting combination represents a valid
probability distribution, we apply a softmax
function to the weighted sum. This ensemble-based approach offers 
advantages: (i) allows our method to use
complementary information from different
views, potentially capturing aspects of the rider's
behavior or environment that may not be visible
from a single view, (ii) the
learnable weighting mechanism provides a
degree of interpretability, as the final weights
can give insights into which of the views are most
informative for the RIP task.

\subsubsection{Implementation Details.}
For data preprocessing, we utilize pre-extracted
VGG-16 features for both single-view and
multi-view tasks. These features are normalized
using z-score normalization. 
Both task methods were trained using AdamW Optimizer, $0.001$ learning rate, weight decay of $1e-5$, batch size of $16$, $20$ epochs, StepLR scheduler with step size=$3$ and gamma=$0.8$, and cross-entropy loss.
We implemented early stopping based on
validation accuracy, saving the best-performing
model with key hyperparameters to be state dimension of the selective layer (D\_state=$32$), kernel size of the convolutional layer in Mamba2 block (D\_conv=$4$), and expansion factor for internal dimension (Expand=$8$).

\begin{figure*}[t]
  \centering
  \includegraphics[width=1\linewidth]{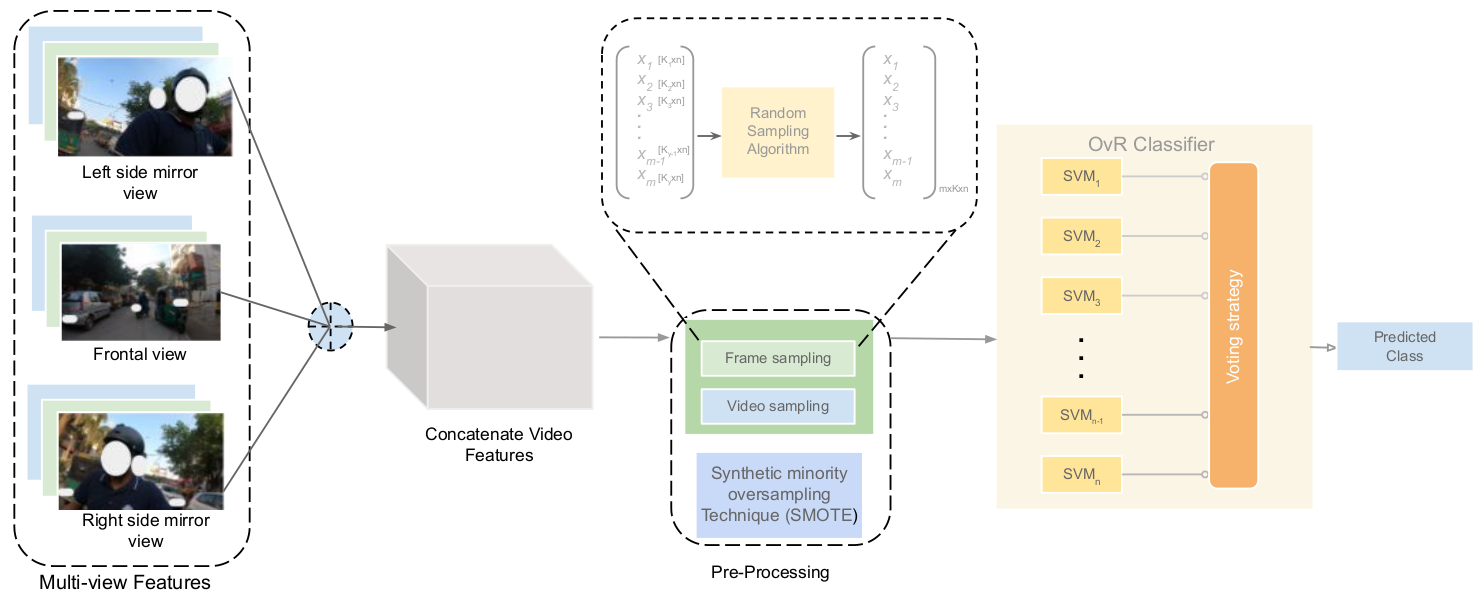}
  \caption{\textbf{SVM based RIP Method.} For frontal and multi-view tasks.}
  \label{fig:svm}\vspace{-5mm}
\end{figure*}

\subsection{\textit{Team ML Cruiser:} SVM based RIP Method}
\label{sec:svm}
This method is contributed by authors Manish Prajapati and Ayesha Choudhary from Jawaharlal Nehru University, New Delhi, and are associated with team \textit{ML Cruiser}.
\subsubsection{Data Preparation.} 
To standardize the dataset, we performed frame-level random sampling to ensure all video features have a uniform number of frames. By randomly selecting frames, the model is exposed to a more diverse set of images, capturing different aspects and nuances in the video. This increases the likelihood of including critical frames that may
be skipped by uniform sampling.

\subsubsection{Synthetic Minority Over-sampling Technique (SMOTE).} 
The dataset exhibits an unbalanced distribution of video features across different classes (see Fig~\ref{fig:data_stats}). To address this
imbalance, we performed oversampling of the minority
class by generating synthetic samples using the SMOTE~\cite{smote}. SMOTE generates synthetic samples for the minority class by interpolating between existing minority class samples and
their nearest neighbors.

\subsubsection{One Vs Rest Classifier.}
We utilize SVM~\cite{svm} for classifying the data as illustrated in Fig.~\ref{fig:svm} for frontal and multi-view RIP tasks because SVMs better handle the high dimensional data and the non-linear relationship between features effectively. Since SVMs are inherently designed for binary classification, we employed a One-vs-Rest
(OvR) strategy with the support vector classifier (SVC) to address the multi-class classification problem.
We trained $K$ binary classifiers $f_K()$ using the SVC with RBF kernel to transform
the training data. This approach allows a non-linear decision surface to be mapped to a linear equation in a higher-dimensional space. The RBF kernel, which is a Gaussian kernel, is particularly useful for transformation when there is
no prior knowledge about the data. To classify a new data point $x$, OvR strategy computes
the decision function values for all $K$ binary classifier $f_K(x)$ and then assigned to the
class with the highest decision support.

\begin{figure*}[t]
  \centering
  \includegraphics[width=1\linewidth]{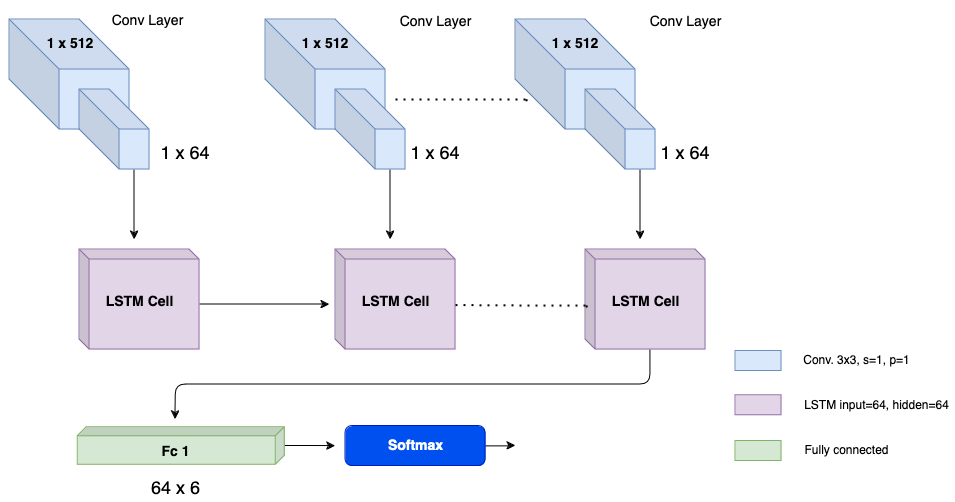}
  \caption{CNN-LSTM based RIP Method.}
  \label{fig:cnnlstm}
\end{figure*}

\subsection{\textit{Team RipTide:} CNN-LSTM based RIP Method}
\label{sec:team3}
This method is contributed by authors Aaryadev Chandra, Dev Chandan, Shireen Chand, and Suvaditya Mukherjee from NMIMS, Mumbai, and are associated with team \textit{RipTide}.
We designed CNN-LSTM method as shown in Fig.~\ref{fig:cnnlstm} to handle the spatial and temporal features for frontal and multi-view. 
The CNN-LSTM for single-view comprised a convolutional block with a 1D conv layer, batch normalization, LeakyReLU activation, and dropout, followed by a bidirectional LSTM with two layers
and a hidden size of $128$. The fully connected layer with a softmax activation generated the final class probabilities. For the multi-view, the CNN-LSTM followed a similar structure of single-view but each view was processed separately through the conv block, concatenated, and passed through the LSTM block, with the output from the last time step passed through the fully connected layer. 

\subsubsection{Implementation Details.}
The best hyperparameters to consider are 
learning rate of $0.001$, batch size of $16$, $400$ epochs, $2$ LSTM layers, $128$ hidden size, $0.25$ dropout rate, and no scheduler. We used Adam optimizer and cross-entropy loss for training.

\begin{table}[h!]
\caption{\textbf{Performance comparison of top submitted results by participating teams.} Accuracy and $F_1$ score (\%) for frontal-view  and multi-view RIP challenges on RAAD dataset.
%"-" indicates that the datasets do not have an in-cabin data source.
}
\centering

\begin{tabular}{c|c|c|c| c|c}
\toprule 
\multirow[c]{2}{*}{\textbf{Team Name}} & \multirow[c]{2}{*}{\textbf{Method}} & \multicolumn{2}{c |}{\textbf{Single-view RIP}} &  \multicolumn{2}{c }{\textbf{Multi-view RIP}} \\

 & & \textbf{Acc.} & \textbf{$F_1$} & 
 \textbf{Acc.} & \textbf{$F_1$} \\
\midrule

Cisco & Mamba2 SSM & \textbf{67.22} & \textbf{66.92} & \textbf{65.22} & \textbf{65.53} \\
ML Cruiser & SVM based RIP & 62.75 & 61.66 & 62.42 & 61.27 \\
RipTide & CNN-LSTM based RIP & 43.29 & 44.15 & 40.74 & 38.65 \\
%CNN LSTM-based (\textit{Team 3 - Resubmis}) & 43.29 & 40.06 & 67.78 & 65.48 \\
\rowcolor{lightgray}
Ours & Baseline & 30.21 & 29.78 & 28.32 & 27.95 \\
Ranesh & - & 24.25 & 20.95 & 18.35 & 15.65 \\
Utkarsh & - & 12.71 & 11.19 & 04.61 & 08.03 \\
\hline

\end{tabular}
\label{tab:baselines}
\vspace{-1cm}
\end{table}
%\vspace{-1cm}

\section{Experimental Results}
%This section presents the experiments carried out by participants in frontal-view and multi-view RIP challenges.

\subsection{Performance Comparison of Various Methods}
\vspace{-.1cm}
Quantitative evaluation results on the RAAD benchmark are summarized in  Table~\ref{tab:baselines} for frontal and multi-view RIP tasks. 
We observe that the Mamba2 method, which is a state-space model outperforms the other two methods and achieves higher accuracy and $F_1$ score. The SVM-based method is 3-5\% lower while CNN-LSTM based method 25\% lower in performance as compared to the best performing method. The \textit{Cisco} team emerged as the winner, excelling in both tasks and across all metrics. The \textit{ML Cruiser} team secured the runner-up position, while \textit{RipTide} achieved third place. Also, we observe that none of the multi-view RIP methods performance improved over frontal-view RIP, with a maximum drop of $2.5\%$ in accuracy and $5.5\%$ in $F_1$ score even after using left and right mirror views. Possible reasons for the superior performance of the frontal view over the multi-view approach include: (i) none of the methods have architecture specifically designed to handle multiple views, and (ii) all methods use the same capacity for training and evaluating multi-view data as for single-view data. Larger networks might yield better scores.  On a positive note, this gap will encourage the research community to build a novel architecture that can better perform on multi-view datasets.
%This could be due to none of these methods exploiting the fusion techniques. ---- better reason for results underperforming for multi-view videos}  

\begin{figure*}[t]
  \centering
  \includegraphics[width=1\linewidth]{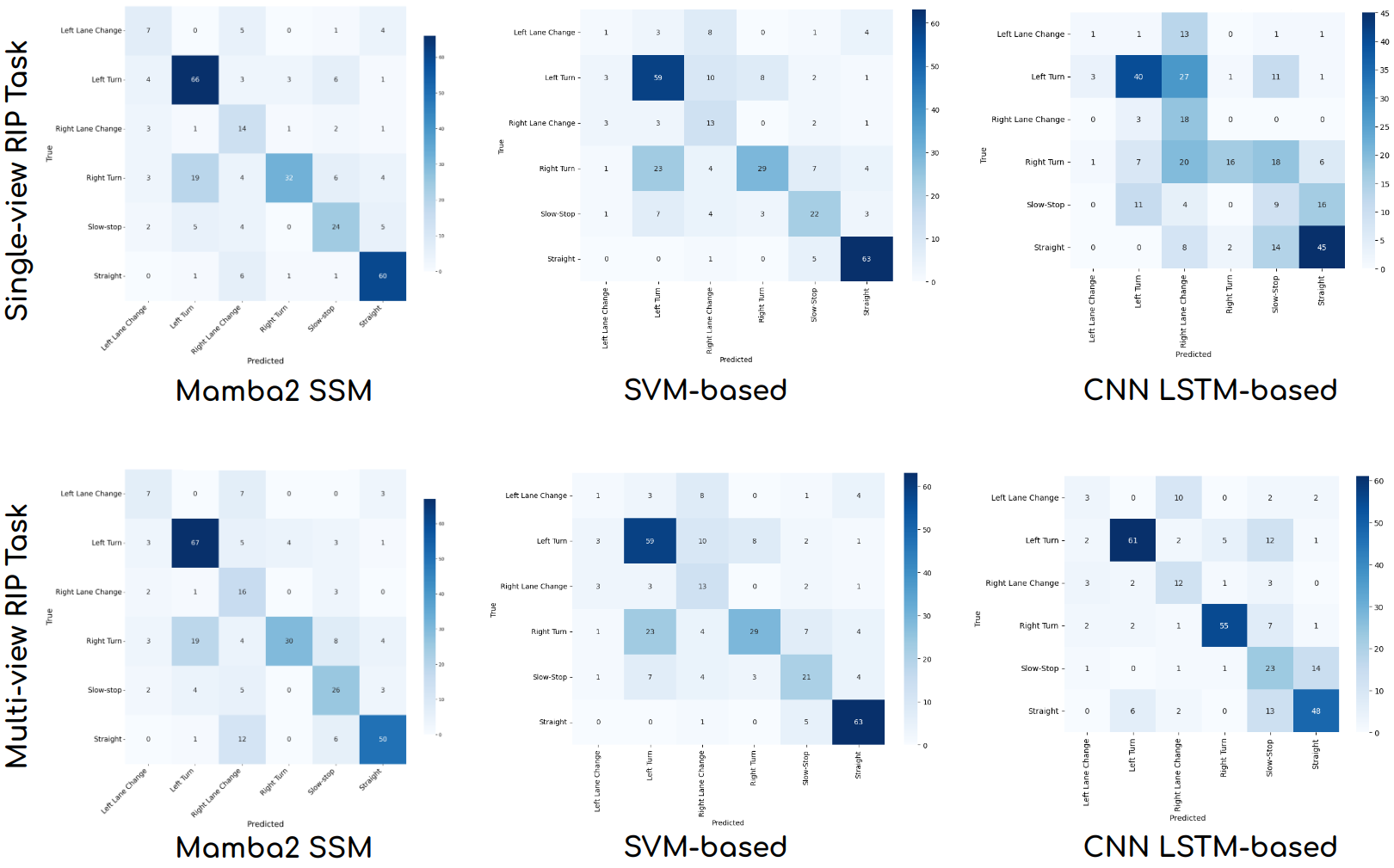}
  \caption{{Confusion matrices of methods for the top three teams.} \textit{(Top row)} frontal-view RIP task and \textit{(Bottom row)} multi-view RIP task on RAAD dataset.}
  \label{fig:confusion}
\end{figure*}
\begin{table}[t]
\caption{\textbf{Performance comparison of maneuver classes for the top three teams.} Various methods accuracy and $F_1$ score (\%) for all the maneuver classes on RAAD dataset.
}
\centering
\begin{tabular}{l|l|l|l |l|l }
 \toprule 
 \multirow[c]{3}{*}{\textbf{Method}} &   \multirow[c]{3}{*}{\textbf{Class}} & \multicolumn{4}{c }{\textbf{Task}}  \\
  & & \multicolumn{2}{c |}{Frontal-view RIP}  & \multicolumn{2}{c }{Multi-view RIP} \\
%  \cline{3-10}
   & & \textbf{Acc. ($\uparrow$)}  & \textbf{$F_1$ ($\uparrow$)} & \textbf{Acc. ($\uparrow$)} & \textbf{$F_1$ ($\uparrow$)} \\ 
   
  %& & \textbf{P ($\uparrow$)} & \textbf{R ($\uparrow$)} & \textbf{Acc. ($\uparrow$)}  & \textbf{$F_1$ ($\uparrow$)} & \textbf{P ($\uparrow$)} & \textbf{R ($\uparrow$)} & \textbf{Acc. ($\uparrow$)} & \textbf{$F_1$ ($\uparrow$)} \\ 
  \midrule
 
 %\multirow{6}{*}{Mamba2 SSM} 
 %& ST & 1 & 0.8000 & 0.8000 & 0.8889 & 1 & 0.8197 &  0.8197 & 0.9009  \\
 %& RT & 0.8649 & 0.8889 & 0.8649 & 0.8767 & 0.8824 & 0.8824 &  0.8824 & 0.8824   \\
 %& LT & 0.7174 & 0.7253 & 0.7174 & 0.7213 & 0.7283 & 0.7363 & 0.7283  & 0.7322 \\
 %& RLC & 0.3889 & 0.4667 & 0.3889 & 0.4242 & 0.3265 & 0.4324 &  0.3265 & 0.3721  \\
 %& LLC & 0.3684 & 0.3684 & 0.3684 & 0.3684 & 0.4118 & 0.4118 &  0.4118 & 0.4118  \\
 %& SS & 0.6000 & 0.6154 & 0.6000 & 0.6076 & 0.5652 & 0.6500 &  0.6047 & 0.5652 \\

 \multirow{6}{*}{Mamba2 SSM} 
 & ST & 80 & 88.89 & 81.97 & 90.09  \\
 & RT & 86.49 & 87.67 & 88.24 & 88.24   \\
 & LT & 71.74 & 72.13 & 72.83  & 73.22 \\
 & RLC & 38.89 & 42.42 & 32.65 & 37.21  \\
 & LLC & 36.84 & 36.84 & 41.18 & 41.18  \\
 & SS & 60 & 60.76 & 60.47 & 56.52 \\
 \bottomrule
 \toprule

% \multirow{6}{*}{SVM based RIP} 
% & ST & 0.4772 & 0.4772 & 0.91 & 0.4772 & 0.4772 & 0.4772 & 0.91 & 0.4772  \\
% & RT & 0.4264 & 0.4027 & 0.43 & 0.4142 & 0.4264 & 0.4027 & 0.43 & 0.4142   \\
% & LT & 0.7108 & 0.7023 & 0.71 & .7065 & 0.7108 & 0.7023 &  0.71 & 0.7065  \\
% & RLC & 0.5652 & 0.5652 & 0.59 & 0.5652 & 0.5652 & 0.5652 & 0.59 & 0.5652  \\
% & LLC & 0.058 & 0.0476 & 0.06 & 0.0522 & 0.058 & 0.0476 &  0.06 & 0.0522  \\
% & SS & 0.4888 & 0.5116 & 0.55 & 0.4999 & 0.4888 & 0.5116 & 0.55 & 0.4999 \\

 \multirow{6}{*}{SVM based RIP} 
 & ST & 91 & 47.72 & 91 & 47.72  \\
 & RT & 43 & 41.42 & 43 & 41.42   \\
 & LT & 71 & 70.65 & 71 & 70.65  \\
 & RLC & 59 & 56.52 & 59 & 56.52  \\
 & LLC & 06 & 5.22 & 6 & 5.22  \\
 & SS & 55 & 49.99 & 55 & 49.99 \\
 
   \bottomrule
 \toprule
 %\multirow{6}{*}{CNN-LSTM based RIP} 
 %& ST & 48.39 & 65.22 & 65.22 & 55.56 & 57.14 & 72.72 &  69.56 & 63.99  \\
 %& RT & 84.21 & 76.19 & 23.53 & 79.99 & 88.71 & 88.71 &  80.88 & 88.7   \\
 %& LT & 64.52 & 64.52 & 48.19 & 64.52 & 85.92 & 79.22 &  73.49 & 82.43  \\
 %& RLC & 20 & 18.37 & 85.72 & 19.15 & 42.86 & 40 &  57.14 & 41.38  \\
 %& LLC & 20 & 20 & 5.88 & 20 & 27.27 & 27.27 &  17.65 & 27.27  \\
 %& SS & 16.98 & 13.43 & 22.5 & 15 & 38.33 & 31.51 &  57.5 & 34.59 \\

  \multirow{6}{*}{CNN-LSTM based RIP} 
 & ST & 65.21 & 65.21 & 69.56 & 71.11  \\
 & RT & 23.52 & 36.78 & 80.88 & 84.61   \\
 & LT & 48.19 & 55.17 & 73.49 & 79.22  \\
 & RLC & 85.71 & 32.43 & 57.14 & 48.97  \\
 & LLC & 5.88 & 9.09 & 17.65 & 21.42  \\
 & SS & 22.5 & 19.35 & 57.5 & 46 \\
 
   \bottomrule
 \toprule

\end{tabular}
\label{Table:2}
\label{tab:benchmarking_table}
\vspace{-0.5cm}
\end{table}

\subsection{Analysis of Confusion Matrices}
In Fig.~\ref{fig:confusion}, confusion matrices are plotted to show
the strengths and weaknesses of each method.
Class imbalance poses a challenge for
learning-based modeling as most methods are
designed with the assumption of an equal number of examples for each class. In RAAD, around $75\%$ videos are labeled as \emph{straight}, \emph{turn} and only $25\%$ labeled as \emph{lane change}, \emph{slow-stop}. Hence, it is important to deal with the class imbalance in these methods. SVM-based method adopts random sampling to avoid the class-imbalanced problem and achieves good results on minority classes. However, CNN-LSTM based method lacks this procedure and fails in the minority classes which indicates the importance of resampling. All these methods specifically confuse between \emph{left turn} and \emph{right turn}. Furthermore, roads with no lane marks lead to wrongly classifying \emph{lane changes} as \emph{straight} and \emph{straight} sometimes as a \emph{slow-stop}.

\subsection{Comparison of Maneuver Classes}
Table~\ref{tab:benchmarking_table} gives the per-class accuracy and $F_1$ scores across the methods on RAAD dataset. We infer that \emph{left lane change} is the most challenging maneuver to be predicted, owing to its low prediction score in all the methods, with $5 - 10\%$ improvement in scores of multi-view RIP task as compared to frontal-view task. Also, prediction scores of \emph{straight} are higher, it can be attributed to many video samples captured in straight. Moreover, no single method gives the best performance on accuracy and $F_1$ score across all maneuver classes. This makes the competition dataset challenging in terms of maneuvers.

\section{Conclusion and Future Direction}
In this competition, we aim to address a vital problem of the two-wheeler rider intention prediction task of anticipating the rider maneuver prior to the actual maneuver in frontal-view and multi-view scenarios. We curate RAAD dataset composed of $1,000$ multi-view video samples with six-rider maneuver classes. We have three different methodologies for the tasks that include one state-space model (Mamba2) and two learning-based methods (SVM and CNN-LSTM). The results indicate that the state-space model outperforms the others overall and maintains balance across maneuver classes. These three approaches provide valuable insights into designing potential solutions for RIP challenges. 

% This competition evaluation contributes to frontal-view and multi-view two-wheeler rider intention prediction tasks for anticipating the rider maneuver prior before the actual maneuver occurs. We curate RAAD dataset composed of $1,000$ multi-view video samples with six rider maneuver classes. We have three different methodologies for the tasks that includes one state-space model (Mamba2) and two learning-based methods (SVM and CNN-LSTM). The results indicate that the state-space model outperforms the others overall and maintains balance across maneuver classes. These three approaches provide valuable insights into designing potential solutions for RIP challenges. 

%However, the tasks addressed in this challenge remain unresolved, highlighting the need for further research in the field of intention prediction.
Future direction of research includes refining and building upon the proposed RIP methods in this competition, inclusion of long-term rider intention videos in the RAAD dataset which would help to minimize the inﬂuence of individual behavioral styles and oﬀer more realistic and valuable insights.
%We also identify the potential paths for further improvements and give insights in future direction of research.

\section*{Acknowledgements}
This work is supported by iHub-Data and Mobility at IIIT Hyderabad.


\end{document}